\documentclass[a4paper,fleqn]{cas-dc}

\renewcommand{\thefootnote}{\fnsymbol{footnote}}
\usepackage[numbers]{natbib}
\usepackage{graphicx}
\usepackage{amsmath,amssymb}
\usepackage{multirow}
\usepackage{booktabs}
\usepackage{float}
\usepackage{gensymb}
\usepackage{newunicodechar}
\usepackage{capt-of}

\newunicodechar{₂}{$_2$}
\newunicodechar{–}{--}
\newunicodechar{−}{-}
\newunicodechar{≤}{$\leq$}
\newunicodechar{≥}{$\geq$}

\begin{document}
\let\printorcid\relax
\let\WriteBookmarks\relax

\shorttitle{Lagrangian-Constrained PPO with KAN for Greenhouse Safe Control}
\shortauthors{Liu and Fan et al.}

\title{Safe Greenhouse Climate Control Using Lagrangian-Constrained PPO with Kolmogorov–Arnold Networks}

\author[1]{Hanzun Liu}
\fnmark[1]

\author[1]{Yuling Fan}
\fnmark[1]

\author[1,2]{Fang Tian}
\author[1]{Zhilong Bie}

\author[1,2]{Zaiwen Feng}
\cormark[1]
\ead{zaiwen.feng@mail.hzau.edu.cn}

\author[3,4]{Yongliang Qiao}
\cormark[1]
\ead{ylqiao@jlu.edu.cn}

\cortext[cor1]{Corresponding author}
\fntext[fn1]{These authors contributed equally to this work.}

\address[1]{College of Informatics, Huazhong Agricultural University, Wuhan 430070, China}
\address[2]{Yunnan Modern Agricultural Industry Research Institute Co., Ltd.,Kunming, Yunnan 650000, China}
\address[3]{College of Biological and Agricultural Engineering, Jilin University, Changchun 130022, China}
\address[4]{Key Laboratory of Efficient Sowing and Harvesting Equipment, Ministry of Agriculture and Rural Affairs, Jilin University, Changchun 130022, China}
\begin{abstract}
Greenhouse climate control requires balancing economic return and maintaining temperature, humidity and CO$_2$ within crop‑adapted growth ranges. Conventional reinforcement learning (RL) greenhouse controllers rely on fixed reward penalties to restrict climate constraint violations, yet such heuristic penalties fail to explicitly constrain long‑term cumulative violations; improperly tuned weights either make policies overly conservative and reduce yields or fail to suppress sustained climate deviations that inhibit photosynthesis and trigger crop diseases. To tackle this limitation, this work first formulates the greenhouse climate regulation task as a Constrained Markov Decision Process (CMDP), and adopts a Lagrangian‑based safe RL framework named Reward Constrained Policy Optimization‑Proximal Policy Optimization (RCPO‑PPO) to separate economic optimization and cumulative safety constraints, which adaptively adjusts penalty intensity without manual weight tuning. To address the strong nonlinear and time‑varying coupling between greenhouse microclimate and crop growth, Kolmogorov–Arnold Networks (KANs) replace standard Multi‑Layer Perceptrons (MLPs) as policy and value approximators for enhanced nonlinear representation, while sinusoidal cyclic time features are embedded into observations to represent diurnal periodic environmental variations. Simulations are carried out on a classic winter lettuce greenhouse dynamic model with 40‑day real‑weather disturbance inputs. Results show that, compared with penalty‑based vanilla Proximal Policy Optimization (PPO), the proposed method reduces cumulative climate constraint violations by 18.65\% and increases the total economic profit of lettuce cultivation by 2.91\%, while stably keeping cumulative climate violations close to the preset safety threshold. These results indicate that the decoupled CMDP‑based constrained optimization together with enhanced KAN‑driven policy representation can effectively mitigate long‑term climate risks while simultaneously improving planting economic benefits, offering a promising constraint‑aware control strategy for precision greenhouse cultivation.

\end{abstract}


\maketitle

\section{Introduction}
Greenhouse plant production enables stable and high-quality food supply by regulating the indoor climate under varying outdoor weather conditions\cite{badji2022design,dsouza2023exploring,vatistas2022systematic}. Key variables such as air temperature, humidity, and carbon dioxide (CO$_2$) concentration strongly influence photosynthesis, biomass accumulation, and plant health\cite{ding2022effect}. In practice, greenhouse operation must simultaneously achieve two long-term goals: (i) maintain the climate within ranges that are suitable for plant growth and disease prevention\cite{tantau2003greenhouse}, and (ii) reduce operational costs related to heating, ventilation, and CO$_2$ enrichment to improve economic returns\cite{rezaei2024optimization,lahlou2025economic}. This creates a fundamental trade-off between production profit and climate regulation reliability.

A wide range of greenhouse climate control methods have been developed, including rule-based strategies\cite{canadas2017improving}, proportional--integral--derivative (PID) controllers\cite{zeng2012nonlinear}, and model predictive control (MPC)\cite{mahmood2023data}. Among them, MPC is attractive because it can explicitly incorporate multivariable dynamics, constraints, and disturbance forecasts\cite{garcia1989model}. However, model-based performance depends on the accuracy of the underlying greenhouse--crop model, and capturing the nonlinear and uncertain interactions between crop physiology, indoor climate, and external disturbances remains challenging\cite{van2010optimal,wang2025global}, especially across different crops, seasons, and operating conditions.

Reinforcement learning has therefore gained increasing attention as a data-driven alternative that can optimize long-horizon objectives through interaction with a simulator or real system\cite{sutton1998reinforcement,zhang2024ai}. Recent studies have shown that Reinforcement learning can improve economic performance in greenhouse control tasks\cite{wang2020deep,xiao2026grower}. Nevertheless, a major barrier to the practical deployment of reinforcement learning in greenhouse climate control is the potentially harmful impact of constraint violations on crop growth and plant health. Prolonged deviations in temperature, humidity, or CO$_2$ concentration can suppress photosynthesis\cite{zheng2019elevated,yusuf2025optimizing,moore2021effect}, promote disease development, and reduce biomass accumulation. In most existing RL-based greenhouse controllers, such climate requirements are enforced only indirectly through penalty terms in the reward function\cite{morcego2023reinforcement,mallick2025reinforcement}. However, designing such heuristic scalar penalty functions is highly non-trivial and often fragile in complex environmental control tasks\cite{tasse2023rosarl}. Although this heuristic approach can reduce violations, it offers no explicit control over the long-horizon magnitude or duration of violations, as these metrics are governed by manually tuned penalty weights. More critically, fixed-penalty formulations frequently lead to suboptimal policies. Heavy penalties imposed for early constraint violations can trap the agent in conservative local optima\cite{mani2025safety}, deterring exploration and severely sacrificing performance, while trivially small penalties fail to deter unsafe behaviors\cite{zhang2022penalized}. Consequently, safety performance fluctuates across different weather conditions and training runs, obscuring the true trade-off between economic profit and crop protection.

These considerations motivate a control framework in which long-horizon constraint violations can be explicitly specified and regulated. In recent years, safe reinforcement learning has emerged as a principled paradigm for handling such requirements and has been successfully applied in safety-critical domains such as robotics\cite{gu2023human}, autonomous driving\cite{wen2020safe}, and energy systems\cite{zhou2022coordinated}, where policies must optimize performance while satisfying strict operational constraints\cite{wachi2020safe,kushwaha2025survey}.

To address these limitations, this study formulate greenhouse climate control as a Constrained Markov Decision Process \cite{altman2021constrained}, where economic profit is optimized subject to explicit constraints on cumulative climate violations. Based on this formulation, we adopt a safe reinforcement learning approach to regulate long-horizon violations during policy learning, rather than relying on heuristic fixed penalties. Beyond explicit constraint regulation, control performance also depends on policy expressiveness and the availability of informative observations. To improve nonlinear representation and account for diurnal dynamics, Proximal Policy Optimization \cite{schulman2017proximal} is augmented with Kolmogorov–Arnold Networks \cite{liu2024kan}, drawing inspiration from recent integration of KANs in online reinforcement learning environments\cite{kich2024kolmogorov}, alongside cyclic time features ($\sin/\cos$ encoding). 

\begin{figure*}[!t]
\centering
\includegraphics[width=0.85\textwidth]{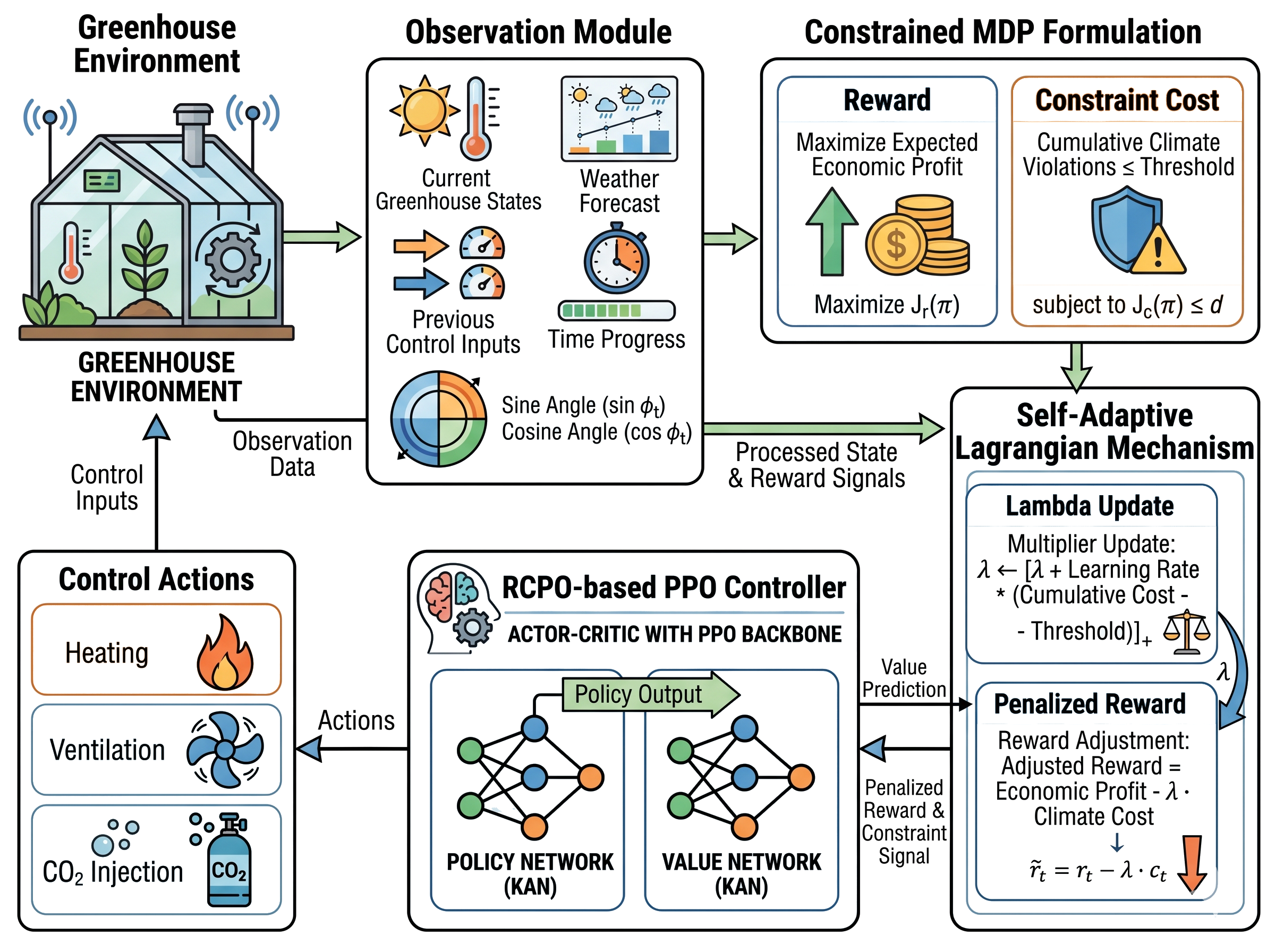}
\caption{Framework of the proposed KAN-based safe reinforcement learning greenhouse climate control system.}
\label{fig:framework}
\end{figure*}

Figure~\ref{fig:framework} illustrates the overall framework of the proposed KAN-based safe reinforcement learning greenhouse climate control system.

In brief, greenhouse climate control is first formulated as a Constrained Markov Decision Process to explicitly distinguish economic objectives and climate regulation constraints. A safe reinforcement learning framework is adopted to suppress cumulative climate violations and attain adjustable trade-offs between profit and long-term safety. Meanwhile, enhanced policy representation together with cyclic time encoding is introduced to improve nonlinear fitting and capture periodic diurnal dynamics.

The main contributions of this work are as follows:
\begin{itemize}
    \item A greenhouse-specific CMDP formulation is developed that explicitly separates economic returns from cumulative climate-risk constraints.
    \item An interpretable cumulative climate-violation cost design is proposed, enabling growers to specify a long-horizon violation budget.
    \item The combined value of adaptive Lagrangian constraint handling, KAN-based policy representation, and cyclic time encoding is systematically validated on a lettuce greenhouse model.
    \item Simulation results show that the integrated controller reduces cumulative climate violations while improving economic profit compared with a penalty-based PPO baseline.
\end{itemize}

\section{Background}

\subsection{Greenhouse Climate Model}
\label{sec:background_model}
\begin{figure*}[!tbp]
    \centering
    \includegraphics[width=0.75\textwidth]{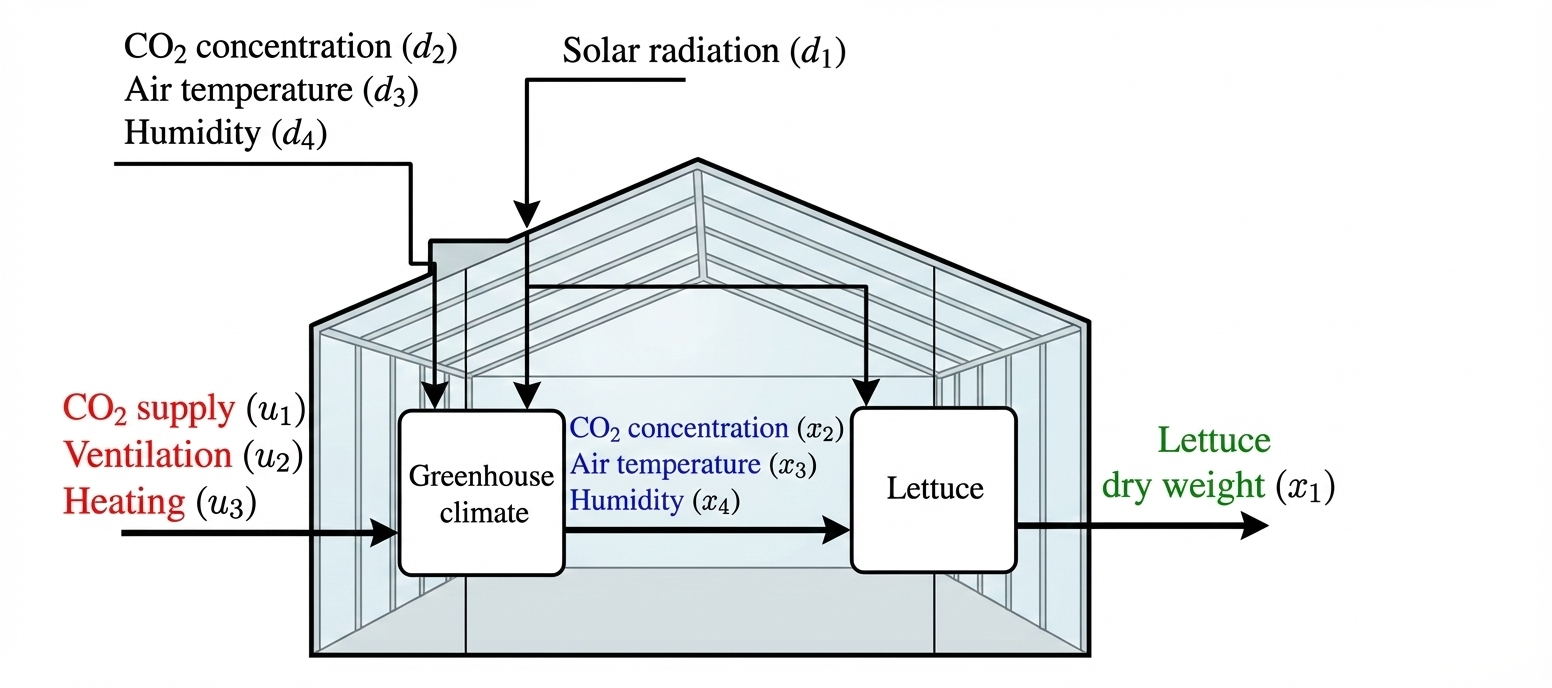}
    \caption{Greenhouse-crop system dynamic relationship diagram.}
    \label{fig:greenhouse}
\end{figure*}

The greenhouse environment considered in this study is a winter lettuce production system described by a nonlinear greenhouse–crop dynamic model proposed in \cite{van1994greenhouse}. This coupled model simultaneously captures indoor microclimate evolution and crop growth dynamics, which is suitable for evaluating climate control strategies related to crop yield and environmental constraints. The structure of the model is shown in Figure \ref{fig:greenhouse}, and the specific definitions of the state, input, output, and disturbance variables are summarized in Table \ref{tab:variables}. 

The continuous-time model is discretized using the fourth-order Runge–Kutta method with a fixed sampling time $\Delta t = 30$ min, resulting in the following discrete-time nonlinear state–space representation:
\begin{equation}
\begin{aligned}
x(k+1) &= f\bigl(x(k),\ u(k),\ d(k),\ p\bigr), \\
y(k)   &= g\bigl(x(k),\ p\bigr).
\end{aligned}
\end{equation}
where $k \in \mathbb{Z}^0$ denotes the discrete-time counter. Specifically, $x(k) \in \mathbb{R}^4$ is the system state vector, $u(k) \in \mathbb{R}^3$ represents the manipulable control input, $d(k) \in \mathbb{R}^4$ accounts for the uncontrollable weather disturbance, and $y(k) \in \mathbb{R}^4$ denotes the system output vector reflecting measurable greenhouse-crop states. The vector $p \in \mathbb{R}^{28}$ stands for the model parameters characterizing the physical, thermal, and biological properties of the system. The explicit expressions of the nonlinear functions $f(\cdot)$ and $g(\cdot)$ are detailed in \ref{app1}.

\begin{table*}[!t]
    \centering
    \caption{Definitions of state, input, output, and disturbance variables}
    \label{tab:variables}
    \small
    \setlength{\tabcolsep}{12pt}
    \begin{tabular}{llll}
    \hline
    \multicolumn{2}{l}{State $x(t)$} & \multicolumn{2}{l}{Output $y(t)$} \\
    $x_1(t)$ & Dry-weight ($\text{kg/m}^2$) & $y_1(t)$ & Dry-weight ($\text{g/m}^2$) \\
    $x_2(t)$ & Indoor $\text{CO}_2$ ($\text{kg/m}^3$) & $y_2(t)$ & Indoor $\text{CO}_2$ (ppm) \\
    $x_3(t)$ & Indoor temperature ($^\circ$C) & $y_3(t)$ & Indoor temperature ($^\circ$C) \\
    $x_4(t)$ & Indoor humidity ($\text{kg/m}^3$) & $y_4(t)$ & Indoor humidity (\%)\\
    \hline
    \multicolumn{2}{l}{Control input $u(t)$} & \multicolumn{2}{l}{Disturbance $d(t)$} \\
    $u_1(t)$ & $\text{CO}_2$ injection ($\text{mg/m}^2/\text{s}$) & $d_1(t)$ & Radiation ($\text{W/m}^2$) \\
    $u_2(t)$ & Ventilation ($\text{mm/s}$) & $d_2(t)$ & Outdoor $\text{CO}_2$ ($\text{kg/m}^3$) \\
    $u_3(t)$ & Heating ($\text{W/m}^2$) & $d_3(t)$ & Outdoor temperature ($^\circ$C) \\
    & & $d_4(t)$ & Outdoor humidity ($\text{kg/m}^3$) \\
    \hline
    \end{tabular}
\end{table*}

\subsection{CMDP and Safety Reinforcement Learning}
\subsubsection{Constrained Markov Decision Process}

A Constrained Markov Decision Process extends the standard Markov Decision Process (MDP) \cite{puterman2014markov} by explicitly incorporating long-term constraints into the decision-making framework. A standard MDP is defined by the tuple $\mathcal{M} = \langle \mathcal{S}, \mathcal{A}, P, r, \gamma \rangle$, where $\mathcal{S}$ is the state space, $\mathcal{A}$ is the action space, $P(s'|s,a)$ is the transition probability, $r(s,a)$ is the reward function, and $\gamma \in (0,1)$ is the discount factor. The classical objective is to find a policy $\pi$ that maximizes the expected discounted return:
\begin{equation}
J_r(\pi) = \mathbb{E}_{\pi}\left[ \sum_{t=0}^{\infty} \gamma^t r(s_t,a_t) \right].
\end{equation}

To introduce safety requirements, a CMDP augments the standard formulation with cost functions and thresholds, defined as $\mathcal{M}_c = \langle \mathcal{S}, \mathcal{A}, P, r, \{c_i\}_{i=1}^m, \gamma, \{d_i\}_{i=1}^m \rangle$. The corresponding optimization problem is formulated as follows:
\begin{subequations}
\label{eq:cmdp_problem}
\begin{align}
\max_{\pi} \quad & J_r(\pi) = \mathbb{E}_{\pi} \left[ \sum_{t=0}^{\infty} \gamma^t r(s_t,a_t) \right], \\
\text{s.t.} \quad & J_{c_i}(\pi) = \mathbb{E}_{\pi} \left[ \sum_{t=0}^{\infty} \gamma^t c_i(s_t,a_t) \right] \le d_i, \nonumber \\
& \quad i=1,\dots,m.
\end{align}
\end{subequations}
Compared with standard MDPs where constraints are implicitly handled via reward penalties, the CMDP framework explicitly separates performance optimization from constraint satisfaction, where the bounds $d_i$ directly quantify the acceptable levels of long-term risk.

\subsubsection{Safe Reinforcement Learning}

Safe Reinforcement Learning aims to learn optimal policies while aiming to satisfy that predefined safety constraints are satisfied during both training and deployment. Guided by the CMDP formulation in \eqref{eq:cmdp_problem}, the constrained optimization problem is typically addressed by introducing Lagrange multipliers $\lambda_i \ge 0$ to construct the following Lagrangian function \cite{altman2021constrained}:
\begin{equation}
\mathcal{L}(\pi,\Lambda) = J_r(\pi) - \sum_{i=1}^m \lambda_i \big(J_{c_i}(\pi)-d_i \big),
\end{equation}
where $\Lambda = \{\lambda_1, \dots, \lambda_m\}$. This formulation enables a primal--dual optimization procedure to balance reward maximization and constraint satisfaction \cite{achiam2017constrained, ray2019benchmarking}. While other prominent Safe RL paradigms, such as Control Barrier Functions (CBFs) or safe shielding, offer step-wise safety filters, they typically demand a highly precise analytical model of system dynamics\cite{kushwaha2025review}. This makes the model-free Lagrangian framework more computationally efficient and naturally aligned with the long-horizon cumulative constraints typical of agricultural environments.

\subsubsection{Reward Constrained Policy Optimization}

Reward Constrained Policy Optimization (RCPO) is a representative Safe RL algorithm that solves the CMDP problem using an online primal--dual scheme \cite{tessler2018reward}. RCPO updates the policy parameters by maximizing the Lagrangian $\mathcal{L}(\pi,\Lambda)$, while the Lagrange multipliers are adaptively updated to penalize constraint violations:
\begin{equation}
\lambda_i \leftarrow \left[ \lambda_i + \alpha_\lambda \big(J_{c_i}(\pi)-d_i \big) \right]_+ , \quad i=1,\dots,m,
\end{equation}
where $\alpha_\lambda$ denotes the learning rate and $[\cdot]_+$ represents the projection onto the nonnegative orthant.

This adaptive mechanism enables RCPO to maintain the cumulative constraint costs close to the predefined thresholds $d_i$, which is particularly suitable for long-term regulation of greenhouse climate violations. 

RCPO exhibits three important properties for greenhouse climate control:
\begin{itemize}
    \item \textbf{Interpretability:} The explicit Lagrange multipliers provide a clear physical meaning for regulating long-term climate violations against the threshold $d$.
    \item \textbf{Adaptability:} The online multiplier update automatically balances economic performance (energy savings) and climate safety under varying weather disturbances.
    \item \textbf{Flexibility:} RCPO is algorithm-agnostic and seamlessly integrates with standard policy gradient methods and advanced deep reinforcement learning frameworks.
\end{itemize}

\subsection{Kolmogorov–Arnold Networks }
Kolmogorov–Arnold Networks have recently been proposed as an alternative neural network architecture inspired by the Kolmogorov–Arnold representation theorem \cite{liu2024kan}. Unlike conventional multilayer perceptrons (MLPs), where nonlinearities are applied at the nodes through fixed activation functions, KANs parameterize nonlinear transformations along the edges using learnable spline functions. This structural difference allows KANs to represent complex nonlinear mappings with a different functional decomposition compared to standard feedforward networks.

In a KAN layer, each connection between neurons is associated with a learnable univariate function, typically implemented via spline-based parameterization. As a result, nonlinear transformations are distributed across edges rather than concentrated at nodes. This formulation provides flexible function approximation while maintaining a relatively compact parameterization. Recent studies suggest that KANs exhibit strong approximation capability and improved interpretability in certain nonlinear regression and control tasks \cite{liu2024kan}. Beyond static regressions, KAN structures have shown remarkable capacity in handling complex time-series forecasting with highly intertwined frequency components\cite{wu2025adakan}, as well as stabilizing policy approximations in continuous online control environments\cite{kich2024kolmogorov}.

Given the highly nonlinear interactions among climate variables, crop physiology, and external disturbances in greenhouse systems, enhanced function approximation capacity is desirable for policy and value estimation. Therefore, KANs are employed as the function approximators within the reinforcement learning framework, replacing standard MLP-based policy and value networks.

\section{Method}

This section details the problem formulation, algorithm implementation, and observation design of the proposed control strategy. The greenhouse climate control task is formulated as a CMDP to explicitly separate economic optimization objectives and long-term climate safety constraints. An RCPO-based safe reinforcement learning framework is adopted to solve the constrained optimization problem under the PPO backbone. Furthermore, KAN-based policy networks and cyclic time feature encoding are introduced to enhance nonlinear fitting ability and capture diurnal greenhouse dynamics. The detailed mathematical formulation and network design are presented below.

\subsection{CMDP formulation for greenhouse climate control}

\paragraph{Control objective.}
Greenhouse climate control aims to maximize economic return while maintaining
indoor climate variables within ranges suitable for crop growth. In practice,
temperature, humidity, and CO$_2$ concentration must be regulated to avoid
conditions that may negatively affect plant development. Therefore, the control
problem involves balancing two objectives: achieving high economic performance
and limiting climate constraint violations over the long growing horizon.
To explicitly represent this trade-off, the greenhouse control problem is
formulated as a Constrained Markov Decision Process.

\paragraph{CMDP components.}
The greenhouse control system is modeled as a CMDP
$(\mathcal{S},\mathcal{A},P,r,c,d,\gamma)$.
At each time step $t$, the agent receives an observation $s_t \in \mathcal{S}$,
selects a control action $a_t \in \mathcal{A}$, and receives a reward $r_t$
together with a constraint cost $c_t$.

The observation vector contains information describing the greenhouse state,
control history, environmental disturbances, and time features. Specifically,
the observation is defined as
\begin{equation}
s_t =
\big[
y(t),\;
u(t-1),\;
w(t:t+N_p),\;
\tau(t),\;
\sin(\phi_t),\;
\cos(\phi_t)
\big],
\end{equation}
where $y(t)$ denotes measurable greenhouse outputs including crop dry weight,
indoor CO$_2$ concentration, air temperature, and relative humidity.
The term $u(t-1)$ represents the previous control input vector.
The variable $w(t:t+N_p)$ denotes the weather look-ahead window: the current values and short-term forecasts of outdoor radiation, CO$_2$ concentration, air temperature, and relative humidity. In this study $N_p = 13$ at the $\Delta t = 30$ min sampling interval, i.e., the current step plus twelve half-hour-ahead steps (a 6 h look-ahead). These values are taken directly from the same measured weather time series that drives the simulator (Section 4.1), i.e., a perfect forecast; the identical sequence is used in training and evaluation, and forecast-error effects are not modeled.
The scalar $\tau(t)$ represents the overall progress of the growing period,
while $\sin(\phi_t)$ and $\cos(\phi_t)$ encode the time-of-day using cyclic
features to capture diurnal dynamics.

The control action corresponds to the greenhouse actuation inputs
\begin{equation}
a_t = u(t) = [u_{CO2}(t), u_{vent}(t), u_{heat}(t)]^\top ,
\end{equation}
which represent CO$_2$ injection, ventilation rate, and heating power,
respectively.

The transition probability $P(s_{t+1}|s_t,a_t)$ is induced by the nonlinear
greenhouse dynamics under exogenous weather disturbances and model
uncertainties.

\paragraph{Economic reward (profit).}
The instantaneous reward reflects the economic profit obtained during one
control interval. It is defined as the crop revenue from incremental biomass
growth minus actuation costs:
\begin{multline}
r_t = c_{DW}\big(x_{DW}(t+1)-x_{DW}(t)\big) \\
- \left(10^{-6}c_{CO2}u_{CO2}(t) + \frac{c_{heat}u_{heat}(t)}{3.6\times10^{6}}\right)\Delta t,
\end{multline}
where $x_{DW}$ is the crop dry weight (kg/m$^2$), $c_{DW}$ is the unit selling
price of the crop dry weight (€/kg), and $c_{CO2}$ and $c_{heat}$ are the unit
costs of CO$_2$ enrichment (€/kg) and heating (€/kWh), respectively.
Here $u_{CO2}$ is in mg/m$^2$/s, $u_{heat}$ in W/m$^2$, and $\Delta t$ in
seconds; the factors $10^{-6}$ (mg$\rightarrow$kg) and $1/(3.6\times10^{6})$
(J$\rightarrow$kWh, since 1 kWh $= 3.6\times10^{6}$ J) make all terms
dimensionally consistent in €/m$^2$. The cumulative profit over the
40-day cultivation horizon is therefore the net return per square meter of
greenhouse floor area, which is the quantity reported in
Table~\ref{tab:performance_comparison}. Ventilation cost is neglected,
consistent with common assumptions in greenhouse climate control studies.

\paragraph{Climate constraint cost (accumulated violation)}
To quantify deviations from acceptable climate conditions, constraint violations
are defined based on the distance between the indoor climate variables and
their allowable ranges.
Let $[y^{min}_{(\cdot)},y^{max}_{(\cdot)}]$ denote the acceptable ranges for
CO$_2$, temperature, and relative humidity.
For each variable, the instantaneous violation is defined as
\begin{equation}
v_{(\cdot)}(t) =
\max\!\big(0,\, y_{(\cdot)}(t)-y^{max}_{(\cdot)}\big)
+
\max\!\big(0,\, y^{min}_{(\cdot)}-y_{(\cdot)}(t)\big),
\end{equation}

The instantaneous constraint cost is then defined as a weighted sum of the
individual violations:
\begin{equation}
c_t =
w_{CO2}\,v_{CO2}(t)
+
w_{T}^{\mathrm{low}}\,v_{T}^{\mathrm{low}}(t)
+
w_{T}^{\mathrm{high}}\,v_{T}^{\mathrm{high}}(t)
+
w_{RH}\,v_{RH}(t),
\end{equation}
where $w_{CO2}$, $w_{T}^{\mathrm{low}}$, $w_{T}^{\mathrm{high}}$, and $w_{RH}$ are fixed normalization constants adopted unchanged from the greenhouse model of van Laatum et al\cite{van2026stochastic}. They map violations of different physical quantities (ppm, °C, \%) onto a comparable scale, with distinct coefficients for lower- and upper-bound temperature violations, and are shared by the baseline and the proposed controller. In fixed-penalty RL the overall penalty scale must be chosen manually; here it is represented by the Lagrange multiplier $\lambda$, which is updated online (Section 3.2).

The expected cumulative violation under policy $\pi$ is
\begin{equation}
J_c(\pi)=
\mathbb{E}_\pi\!\left[
\sum_{t=0}^{T-1}
\gamma^t c_t
\right],
\end{equation}
which aggregates both the magnitude and duration of constraint violations.

\paragraph{CMDP objective.}
The CMDP objective is to maximize the expected cumulative reward while
limiting the cumulative violation:
\begin{equation}
\begin{aligned}
\max_{\pi} \quad
& J_r(\pi)=
\mathbb{E}_\pi
\left[
\sum_{t=0}^{T-1}
\gamma^t r_t
\right], \\
\text{s.t.}\quad
& J_c(\pi) \le d ,
\end{aligned}
\end{equation}
where $d$ represents a user-defined violation budget that limits the allowable
long-horizon climate deviation.

\subsection{RCPO-based controller implementation}
In this work, we instantiate the CMDP with a single aggregate constraint cost $J_c = \mathbb{E}[\sum_t \gamma^t c_t] \le d$, so a single multiplier $\lambda$ is maintained.

The CMDP formulation introduced above is solved using Reward Constrained
Policy Optimization, with Proximal Policy Optimization serving
as the policy optimization backbone. RCPO converts the constrained
optimization problem into a Lagrangian formulation while maintaining
compatibility with standard actor–critic reinforcement learning algorithms.
This makes it particularly convenient to integrate with PPO without requiring
major modifications to the underlying policy optimization framework.

Specifically, the CMDP objective is transformed into the following Lagrangian
function:
\begin{equation}
\mathcal{L}(\pi,\lambda)=J_r(\pi)-\lambda\big(J_c(\pi)-d\big),
\end{equation}
where $J_r(\pi)$ and $J_c(\pi)$ denote the expected cumulative reward and
constraint cost under policy $\pi$, respectively. The nonnegative multiplier
$\lambda \ge 0$ balances the trade-off between economic profit and climate
regulation safety.

In practice, optimizing the Lagrangian objective is equivalent to using a
penalized reward signal
\begin{equation}
\tilde r_t = r_t - \lambda c_t ,
\end{equation}
which dynamically penalizes climate constraint violations according to the
current multiplier value. In this formulation, the actor and critic are trained
using the penalized reward signal, while the constraint evaluation used to
update the multiplier is computed from the original constraint cost.

During training, the policy parameters are optimized using PPO to maximize
the Lagrangian objective through policy gradient updates based on the modified
reward signal $\tilde r_t$. Meanwhile, the Lagrange multiplier is updated every $K=4$ training episodes according to
\begin{equation}
\lambda \leftarrow
\big[\lambda+\alpha_\lambda(J_c(\pi)-d)\big]_+ ,
\end{equation}
where $\alpha_\lambda$ denotes the multiplier learning rate and $[\cdot]_+$
represents projection onto the nonnegative orthant.

This procedure can be interpreted as a primal–dual optimization process in
which the policy parameters are updated to maximize the Lagrangian objective,
while the multiplier adapts to enforce the cumulative violation constraint.
Such an adaptive penalty mechanism avoids the need for manually tuning fixed
penalty coefficients and enables explicit regulation of long-horizon climate
violations.

\subsection{Policy Representation and Observation Design}
\subsubsection{KAN-based policy and value networks}

In standard implementations of PPO, the policy and value functions are
typically parameterized using multi-layer perceptrons (MLPs).
Although MLPs are widely used due to their simplicity and general
approximation capability, their node-based activation structure may
limit representation efficiency when modeling highly nonlinear control
dynamics such as those present in greenhouse climate systems.

To improve the expressiveness of the policy representation, the
MLP-based networks in PPO are replaced with Kolmogorov–Arnold Networks
. KAN is a recently proposed neural network architecture in which
nonlinear transformations are modeled by learnable functions on network
edges rather than fixed activation functions on nodes.
According to the Kolmogorov–Arnold representation theorem, multivariate
functions can be decomposed into compositions of univariate functions,
which provides the theoretical basis for the KAN architecture
\cite{liu2024kan}.

Within the PPO framework, the overall actor–critic structure remains
unchanged. The observation vector $s_t$ is directly fed into the neural
network, and the policy network outputs the action distribution
$\pi(a_t|s_t)$ while the value network estimates the state value
$V(s_t)$. The difference from the standard PPO implementation lies in
the function approximator: the conventional MLP networks used for the
policy and value functions are replaced by KAN networks with identical
input and output interfaces.

This replacement allows the reinforcement learning agent to utilize
KAN's flexible nonlinear function representation while preserving the
original PPO training procedure. Consequently, the policy optimization,
value estimation, and RCPO-based constraint handling remain unchanged,
while the expressive capacity of the neural function approximator is
enhanced.

To facilitate reproduction, Table~\ref{tab:kan_settings} summarizes the
network architecture and implementation settings of the KAN-based policy
and value networks.

\begin{table}[t]
\centering
\caption{Network and implementation settings of the KAN-based actor and critic networks.}
\label{tab:kan_settings}
\begin{tabular}{ll}
\toprule
Item & Value \\
\midrule
Observation dimension $n_{\mathrm{obs}}$ & $62$ \\
KAN layers (actor) & 3: $[62,\,128,\,128,\,128]$ \\
KAN layers (critic) & 3: $[62,\,128,\,128,\,128]$ \\
Spline order $k$ & 3 \\
Grid size & 3 \\
Grid range & $[-1,\,1]$ \\
Base function & SiLU \\
Actor/critic hidden width & Identical ($128$) \\
Optimizer & Adam \\
Learning rate & $1\times10^{-4}$ \\
\bottomrule
\end{tabular}
\end{table}

\subsubsection{Time feature encoding}

Greenhouse climate dynamics exhibit strong diurnal patterns due to
periodic variations in solar radiation, outdoor temperature, and
operational schedules. To allow the policy to account for these
time-dependent dynamics, explicit time information is incorporated
into the observation vector.

Specifically, the time of day is encoded using sinusoidal functions
to represent its cyclic nature:
\begin{equation}
t_{sin} = \sin\left(\frac{2\pi t}{T_{day}}\right), \qquad
t_{cos} = \cos\left(\frac{2\pi t}{T_{day}}\right),
\end{equation}
where $t$ denotes the current time step and $T_{day}$ corresponds to
the length of one day in the simulation.

The resulting features $(t_{sin}, t_{cos})$ are appended to the
observation vector and provided as additional inputs to the policy
and value networks. This cyclic encoding enables the reinforcement
learning agent to capture periodic environmental dynamics while
avoiding discontinuities associated with raw time representations.

\section{Experimental Setup}

\subsection{Simulation environment}

All experiments were conducted in simulation using a nonlinear greenhouse--crop
dynamic model for winter lettuce production originally proposed by
\cite{van1994greenhouse}. The model describes the coupled dynamics between crop
biomass accumulation and greenhouse climate, including indoor
\(\mathrm{CO}_2\) concentration, air temperature, and humidity, under the
influence of control actions and exogenous weather disturbances.

To account for model mismatch and biological variability, parametric uncertainty
was explicitly introduced by perturbing the nominal model parameters with a
relative uncertainty level of \(2.5\%\). Specifically, Parametric uncertainty was applied to all 28 model parameters; at every control step each parameter was resampled independently from a uniform distribution around its nominal value with a total relative range of  \(\pm 2.5\%\), using a seeded RNG.

The continuous-time greenhouse model was discretized using a fourth-order
Runge--Kutta method with a fixed sampling time of \(\Delta t = 30\) min,
resulting in a discrete-time nonlinear state--space system used for both
controller training and evaluation.

The reinforcement learning agent observes the state vector defined in
Section~3, which includes greenhouse measurements (crop dry-weight,
indoor \(\mathrm{CO}_2\) concentration, indoor air temperature, and relative
humidity), previous control inputs, weather forecasts, and time features.
The agent applies three control inputs: \(\mathrm{CO}_2\) injection,
ventilation rate, and heating power.

\paragraph{Economic parameters and constraint coefficients.}

The economic reward and climate-violation cost are parameterized using the
coefficients summarized in Table~\ref{tab:econ_params}. These include crop
price, energy costs, and weighting coefficients used to quantify cumulative
constraint violations.

\begin{table*}[!t]
\centering
\caption{Economic parameters and constraint cost coefficients.}
\label{tab:econ_params}
\begin{tabular}{lccc}
\hline
Variable & Value & Unit & Description \\
\hline
$c_{CO_2}$ & 0.1906 & €/kg & CO$_2$ price coefficient \\
$c_{heat}$ & 0.1281 & €/kWh & Heating price coefficient \\
$c_{DW}$ & 22.29 & €/kg & Crop dry-weight price \\
$\lambda_{CO_2}$ & $5\times10^{-5}$ & -- & Weight for CO$_2$ violations \\
$\lambda_{T}^{\min}$ & $3\times10^{-3}$ & -- & Weight for lower temperature violations \\
$\lambda_{T}^{\max}$ & $5\times10^{-3}$ & -- & Weight for upper temperature violations \\
$\lambda_{RH}$ & $7\times10^{-4}$ & -- & Weight for humidity violations \\
\hline
\end{tabular}
\end{table*}

\paragraph{Climate and control constraints.}

The acceptable operating ranges for climate variables and the allowable control
input ranges are summarized in Table~\ref{tab:constraints}. These bounds define
the target crop-suitable operating region and are used to compute the cumulative
violation cost in the CMDP formulation.

\begin{table}[t]
\centering
\caption{Climate variable ranges and control input limits.}
\label{tab:constraints}
\begin{tabular}{lccc}
\hline
Variable & Lower & Upper & Unit \\
\hline
\multicolumn{4}{l}{\textit{Climate variables}} \\
$y_{CO_2}$ & 500 & 1600 & ppm \\
$y_T$ & 10 & 20 & $^\circ$C \\
$y_{RH}$ & 0 & 80 & \% \\
\hline
\multicolumn{4}{l}{\textit{Control inputs}} \\
$u_{CO_2}$ & 0 & 1.2 & mg/m$^2$/s \\
$u_{heat}$ & 0 & 150 & W/m$^2$ \\
$u_{vent}$ & 0 & 7.5 & m$$/m$$/s \\
\hline
\end{tabular}
\end{table}

Weather disturbances were provided to the simulator as exogenous time-series
inputs, including global solar radiation, outdoor \(\mathrm{CO}_2\) concentration,
outdoor air temperature, and outdoor humidity. The weather data were taken from
real-world measurements recorded at Bleiswijk, The Netherlands, starting on
February 9, 2014, and covering a continuous cultivation period of 40 days.

The nominal weather data were the same for all controllers. Nevertheless,
as they represent exogenous disturbances, white noise with a standard
deviation of \(0.3\) was added to generate a different perturbed weather
profile in each training episode and evaluation run, thereby exposing the
policies to a wider range of weather conditions. The nominal data and the
noise level were identical for all controllers to ensure a fair
comparison \cite{morcego2023reinforcement}.

\subsection{Training and hyperparameter settings}

Proximal Policy Optimization was adopted as the policy optimization
backbone for all reinforcement learning controllers, with the underlying 
algorithms and training workflows implemented utilizing the Stable-Baselines3 
framework~\cite{raffin2021stable}. The PPO hyperparameters
were kept identical across experiments and are summarized in
Table~\ref{tab:ppo_hyperparams}.

\begin{table}[H]
\centering
\caption{PPO hyperparameters used in the experiments.}
\label{tab:ppo_hyperparams}
\begin{tabular}{lc}
\hline
Hyperparameter & Value \\
\hline
Total training timesteps & $4 \times 10^6$ \\
Number of parallel environments ($n_{\mathrm{envs}}$) & $8$ \\
Rollout length ($n_{\mathrm{steps}}$) & $1920$ \\
Batch size & $1920$ \\
Optimization epochs per update & $10$ \\
Discount factor ($\gamma$) & $0.98$ \\
GAE parameter ($\lambda_{\mathrm{GAE}}$) & $0.95$ \\
Clipping range & $0.1$ \\
Entropy coefficient & $0.01$ \\
Value function coefficient & $0.5$ \\
Maximum gradient norm & $0.5$ \\
\hline
\end{tabular}
\end{table}

To enforce the long-horizon climate constraint, PPO was augmented with
Reward Constrained Policy Optimization, which introduces a Lagrange
multiplier to regulate the trade-off between economic performance and
cumulative climate-violation cost. The RCPO-specific settings are listed in
Table~\ref{tab:rcpo_hyperparams}.

\begin{table}[H]
\centering
\caption{RCPO hyperparameters for constraint regulation.}
\label{tab:rcpo_hyperparams}
\begin{tabular}{lcc}
\hline
Hyperparameter & Symbol & Value \\
\hline
Initial penalty coefficient & $\lambda_0$ & $1.0$ \\
Minimum penalty coefficient & $\lambda_{\min}$ & $0.05$ \\
Maximum penalty coefficient & $\lambda_{\max}$ & $50.0$ \\
Constraint threshold & $d$ & $0.7$ \\
Penalty update interval (episodes) & $K$ & $4$ \\
Penalty learning rate & $\alpha_{\lambda}$ & $0.01$ \\
\hline
\end{tabular}
\end{table}

\section{Results and Discussion}

\subsection{Main performance comparison}

The closed-loop performance of the proposed RCPO-PPO controller is
quantitatively compared with the baseline penalty-based PPO controller.
Economic performance is evaluated using cumulative profit, while safety
performance is measured by the cumulative climate-violation cost defined in
Section~3. Both controllers are evaluated under identical greenhouse
dynamics, weather disturbances, and training settings.

\begin{table*}[!t]
\centering
\caption{Performance comparison under different controller configurations.}
\label{tab:performance_comparison}
\begin{tabular}{lcccc}
\hline
Method & Cumulative profit & $\Delta$ Profit (\%) & Cumulative violation & $\Delta$ Violation (\%) \\
\hline
PPO (baseline)           & $3.782$ ($\pm0.146$) & -- ($\pm3.86$)        & $0.858$ ($\pm0.027$) & -- ($\pm3.15$)        \\
\hline
RCPO w/o KAN \& Time     & $3.543$ ($\pm0.061$) & $-6.32$ ($\pm1.72$)   & $0.695$ ($\pm0.026$) & $-19.00$ ($\pm3.74$)   \\
RCPO w/o Time            & $3.635$ ($\pm0.064$) & $-3.89$ ($\pm1.76$)   & $0.712$ ($\pm0.039$) & $-17.02$ ($\pm5.48$)   \\
RCPO w/o KAN             & $3.755$ ($\pm0.165$) & $-0.71$ ($\pm4.39$)   & $0.710$ ($\pm0.025$) & $-17.25$ ($\pm3.52$)   \\
KAN-RCPO-PPO             & $3.892$ ($\pm0.147$) & $+2.91$ ($\pm3.78$)   & $0.698$ ($\pm0.029$) & $-18.65$ ($\pm4.15$)   \\
\hline
\end{tabular}
\end{table*}

As shown in Table~\ref{tab:performance_comparison}, the proposed RCPO-PPO
controller achieves a clear reduction in cumulative climate violations while
improving economic performance. Compared with the baseline PPO controller,
the cumulative violation cost is reduced by approximately $18.65\%$,
indicating a substantially improved ability to regulate long-horizon
climate constraint violations.

At the same time, the cumulative profit increases by approximately
$2.91\%$, indicating that improved safety performance is achieved without
sacrificing economic return. These results indicate that the CMDP-based
formulation allows the controller to regulate cumulative violations near
the predefined threshold rather than relying on heuristic penalty tuning.

Taken together, the experimental results show that the proposed approach
successfully achieves the desired trade-off between economic performance
and climate regulation safety: cumulative violations are effectively
regulated near the specified limit, while economic profit is improved.

\subsection{Ablation study on policy representation enhancements}

To evaluate the contribution of the proposed representation improvements,
an ablation study was conducted before introducing the RCPO safety
constraint mechanism. In this experiment, three controller configurations
were compared under identical training conditions: the baseline PPO
controller, PPO augmented with cyclic time features, and PPO using both
time features and KAN-based policy networks.

\begin{figure}
\centering
\includegraphics[width=.9\columnwidth]{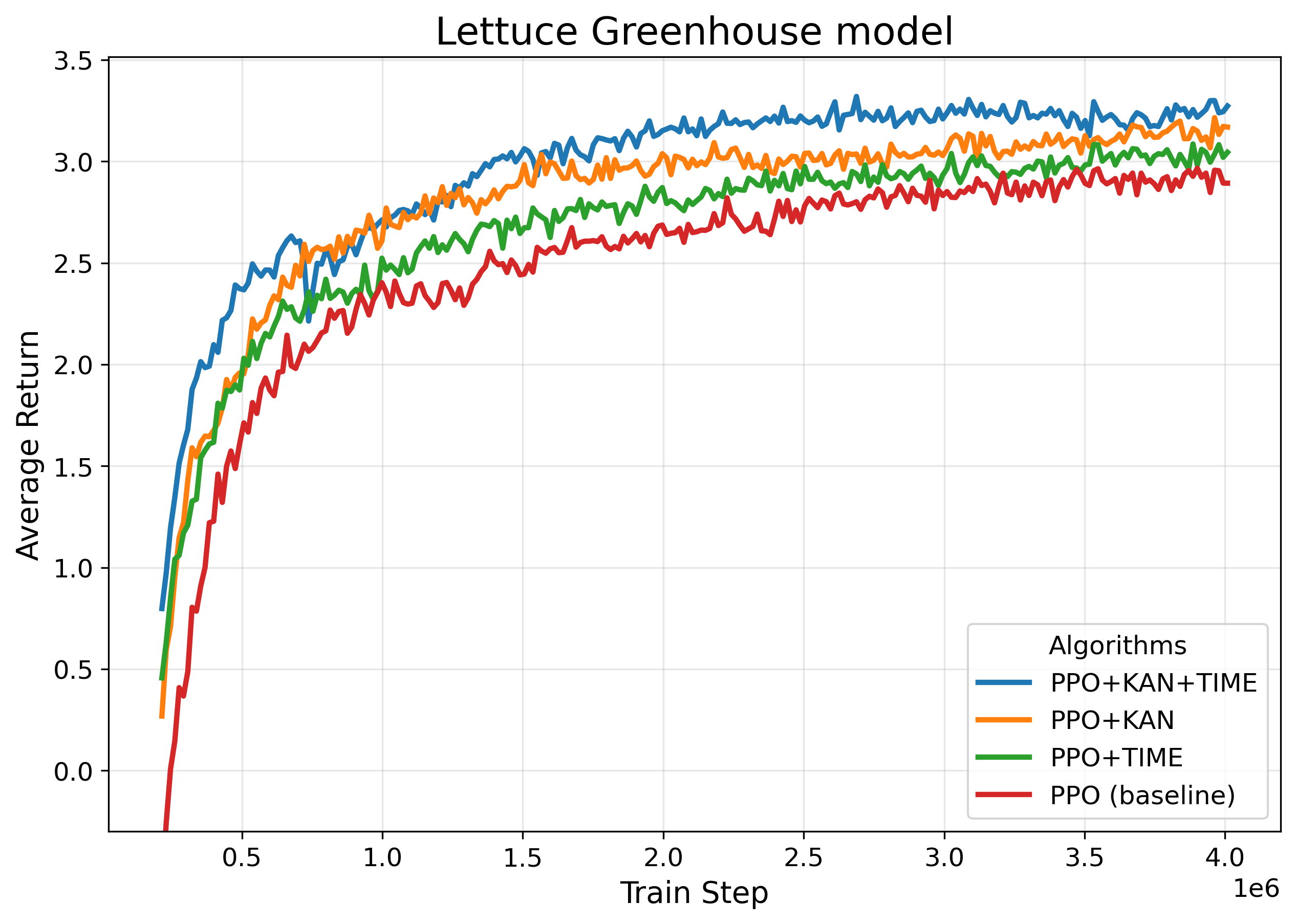}
\caption{Ablation study on policy representation enhancements. Training
performance comparison between baseline PPO, PPO with cyclic time
features, and PPO with KAN-based policy networks. All methods are
trained without RCPO constraint regulation.}
\label{fig:ablation}
\end{figure}

All controllers were trained using the same reward formulation without
constraint regulation, allowing the impact of the proposed representation
enhancements to be isolated from the safe reinforcement learning component.
The average episodic return during training is shown in
Fig.~\ref{fig:ablation}.

As illustrated in Fig.~\ref{fig:ablation}, incorporating cyclic time
features improves the training performance compared with the baseline
PPO controller. This improvement is attributed to the ability of the
time encoding to capture the strong diurnal patterns present in
greenhouse climate dynamics.

Furthermore, replacing the conventional MLP policy network with a
KAN-based architecture leads to additional performance gains. The
KAN-based controller achieves both faster convergence during the
early training stage and a higher final average return.

Overall, the results demonstrate that the proposed representation
enhancements—time feature encoding and KAN-based policy
approximation—significantly improve the learning capability of the
reinforcement learning controller. These improvements provide a
stronger policy representation foundation before introducing the
RCPO-based safety constraint mechanism.

\subsection{Reduction in climate constraint violations by variable}

\begin{figure}
\centering
\includegraphics[width=.9\columnwidth]{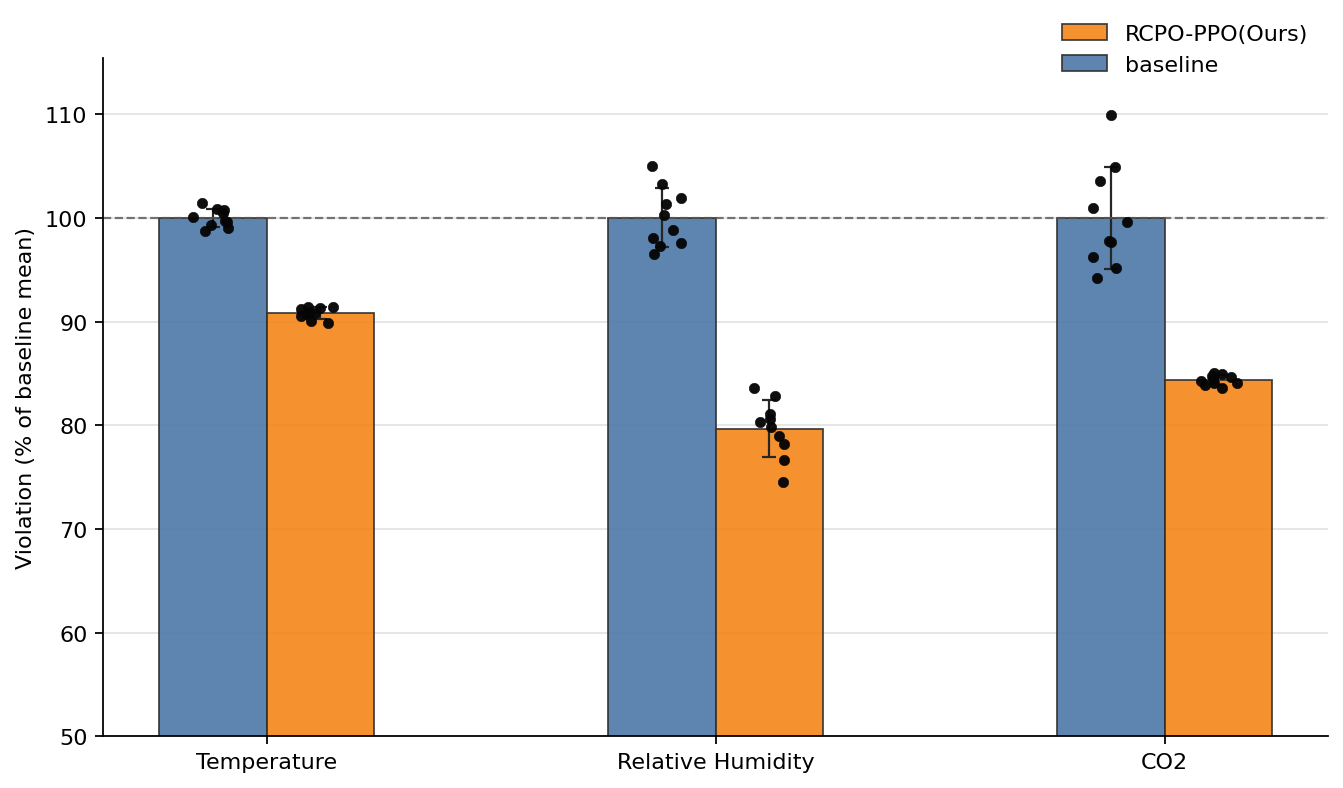}
\caption{Normalized cumulative climate constraint violations for different
climate variables achieved by the proposed RCPO-PPO controller relative to
the baseline PPO controller. The baseline PPO result is normalized to
\(100\%\).}
\label{fig:reduce_violation_components}
\end{figure}

To further examine how the proposed controller improves safety performance,
the cumulative constraint violations are decomposed into three climate
variables: relative humidity, air temperature, and indoor \(\mathrm{CO}_2\)
concentration. The normalized cumulative violations achieved by the proposed
RCPO-PPO controller relative to the baseline PPO controller are shown in
Fig.~\ref{fig:reduce_violation_components}.

As illustrated in Fig.~\ref{fig:reduce_violation_components}, the largest
improvement is observed in humidity regulation, where the cumulative
violation is reduced by \(20.3\%\). The cumulative
\(\mathrm{CO}_2\) violation is reduced by \(15.6\%\), while the cumulative
temperature violation is reduced by \(9.2\%\). These results indicate that
the proposed safe reinforcement learning framework consistently reduces
constraint violations across all climate variables, with the strongest
improvement observed for relative humidity.

From a practical greenhouse management perspective, excessive humidity is associated with conditions conducive to condensation and disease. Therefore, the observed improvement suggests that the proposed controller may reduce conditions associated with such risks, but disease incidence itself is not modeled in this study.

\subsection{Reduction in heating and \texorpdfstring{\(\mathrm{CO}_2\)}{CO2} input usage}

\begin{figure}
\centering
\includegraphics[width=.9\columnwidth]{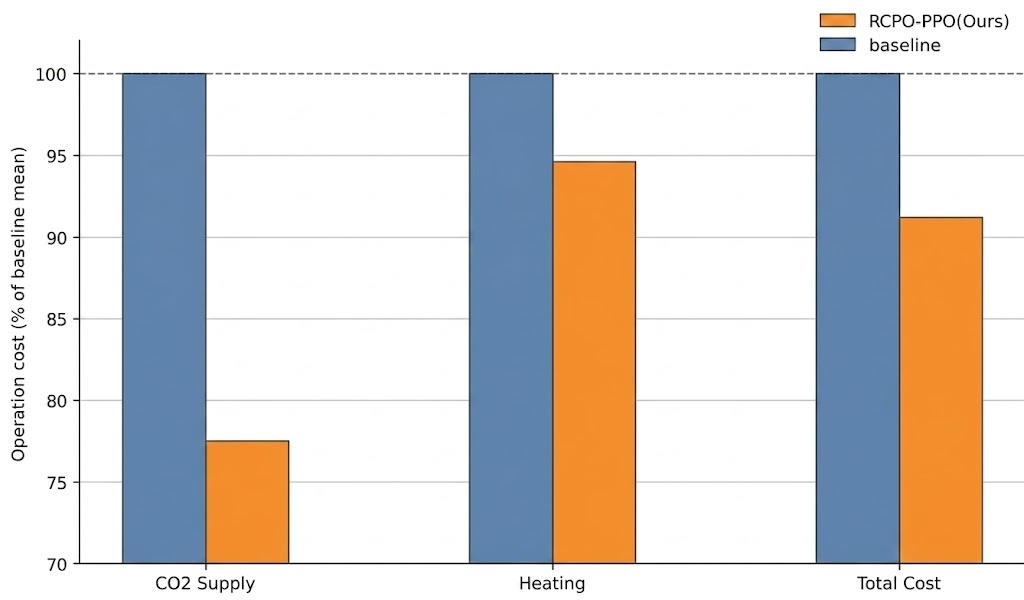}
\caption{Normalized cumulative operation costs associated with heating and
\(\mathrm{CO}_2\) input achieved by the proposed RCPO-PPO controller
relative to the baseline PPO controller. The baseline PPO result is
normalized to \(100\%\).}
\label{fig:reduce_energy_usage}
\end{figure}

Besides improving climate regulation safety, the proposed controller also
affects resource consumption. To evaluate this aspect, the cumulative
operation costs associated with heating and \(\mathrm{CO}_2\) input are
compared with those of the baseline PPO controller. The corresponding
normalized costs are shown in Fig.~\ref{fig:reduce_energy_usage}.

As shown in Fig.~\ref{fig:reduce_energy_usage}, the proposed controller
reduces the cumulative \(\mathrm{CO}_2\) input cost by \(22.5\%\) and the
cumulative heating cost by \(5.4\%\) relative to the baseline PPO
controller. When both cost components are combined, the total operation cost
is reduced by \(8.8\%\). These reductions indicate that improved safety
performance is not achieved through excessive actuator usage. Instead, the
controller is able to regulate the greenhouse climate more efficiently while
lowering resource consumption.

This trend aligns well with the profit improvement reported in
Table~\ref{tab:performance_comparison}. Since heating and
\(\mathrm{CO}_2\) enrichment are the main economic costs in the reward
function, their reduced usage directly contributes to improved economic
performance. Overall, the proposed method achieves a favorable balance
between climate safety, control efficiency, and operational profitability.

\section{Conclusion and Future Work}

This work develops a safe reinforcement learning framework for greenhouse climate control to explicitly balance economic performance and long-term climate regulation safety. The greenhouse control task is cast into a Constrained Markov Decision Process formulation with explicit cumulative constraints imposed on cumulative climate violations.

To solve the resulting CMDP, a Reward Constrained Policy Optimization approach is adopted, enabling direct regulation of long-horizon constraint violations through an adaptive Lagrange multiplier. In addition, the policy representation is enhanced by incorporating Kolmogorov–Arnold Networks and cyclic time features, which improve the ability of the controller to capture nonlinear greenhouse dynamics and diurnal patterns.

Simulation results based on a standard lettuce greenhouse model demonstrate that the proposed method effectively reduces cumulative climate violations by approximately \(18.65\%\) while increasing economic profit by about \(2.91\%\) compared with a penalty-based PPO baseline. Further analysis shows that RCPO is primarily responsible for enforcing constraint satisfaction, whereas KAN-based policy networks and time feature encoding contribute to improved economic performance. Additional results on individual climate variables and resource usage indicate that the proposed controller achieves significant reductions in humidity-related violations and decreases heating and \(\mathrm{CO}_2\) input usage, realizing efficient and reliable greenhouse operation with a controllable trade-off between profitability and operational safety.

Future work will focus on extending the proposed method to more diverse crop types and seasonal conditions, as well as validating the approach in real-world greenhouse environments. In addition, integrating more advanced safe reinforcement learning algorithms and incorporating uncertainty-aware decision-making may further improve robustness and practical applicability.

\appendix
\renewcommand{\thesection}{Appendix \Alph{section}}
\section{Lettuce Greenhouse Model}\label{app1}

This appendix provides the continuous-time nonlinear lettuce greenhouse model
used in this study. The model is adapted from \cite{van1994greenhouse} and is
consistent with the state, input, output, and disturbance definitions introduced
in Section~\ref{sec:background_model}. The discrete-time model used in controller
training and evaluation is obtained by discretizing the following continuous-time
system with the fourth-order Runge--Kutta method using a sampling time
$\Delta t = 30$ min.

\subsection*{State, input, output, and disturbance variables}

The state, input, disturbance, and output vectors are defined as
\begin{equation}
\begin{aligned}
x(t) &= [x_1(t),x_2(t),x_3(t),x_4(t)]^\top,\\
u(t) &= [u_1(t),u_2(t),u_3(t)]^\top,\\
d(t) &= [d_1(t),d_2(t),d_3(t),d_4(t)]^\top,\\
y(t) &= [y_1(t),y_2(t),y_3(t),y_4(t)]^\top .
\end{aligned}
\end{equation}

Here, $x(t)$ denotes the greenhouse--crop state vector, $u(t)$ denotes the
control input vector, $d(t)$ denotes the weather disturbance vector, and
$y(t)$ denotes the measurable output vector. The detailed physical meanings
of these variables are given in Table~\ref{tab:variables}.

\subsection*{Continuous-time dynamics}

The continuous-time greenhouse model is given by
\begin{equation}
\dot{x}_1(t)
=
p_1 \phi_{\mathrm{phot},c}(t)
-
p_2 x_1(t)\, 2^{x_3(t)/10 - 5/2},
\end{equation}

\begin{equation}
\begin{split}
\dot{x}_2(t)
=
\frac{1}{p_9}
\Big(
&-\phi_{\mathrm{phot},c}(t)
+
p_{10} x_1(t)\, 2^{x_3(t)/10 - 5/2} \\
&+
10^{-6}u_1(t)
-
\phi_{\mathrm{vent},c}(t)
\Big),
\end{split}
\end{equation}

\begin{equation}
\begin{split}
\dot{x}_3(t)
=
\frac{1}{p_{16}}
\Big(
&u_3(t)
-
\left(10^{-3}p_{17}u_2(t)+p_{18}\right) \\
&\times
\bigl(x_3(t)-d_3(t)\bigr)
+
p_{19}d_1(t)
\Big),
\end{split}
\end{equation}

\begin{equation}
\dot{x}_4(t)
=
\frac{1}{p_{20}}
\left(
\phi_{\mathrm{transp},h}(t)
-
\phi_{\mathrm{vent},h}(t)
\right).
\end{equation}

\subsection*{Measurement equations}

The measurable outputs are defined as
\begin{equation}
y_1(t)=10^3 x_1(t),
\end{equation}

\begin{equation}
y_2(t)=
10^3 x_2(t)\,
\frac{p_{12}\bigl(x_3(t)+p_{13}\bigr)}{p_{14}p_{15}},
\end{equation}

\begin{equation}
y_3(t)=x_3(t),
\end{equation}

\begin{equation}
\begin{split}
y_4(t)
=
\frac{10^2}{11}\,
x_4(t)\,
p_{12}\bigl(x_3(t)+p_{13}\bigr)
\exp\left(
\frac{p_{27}x_3(t)}{x_3(t)+p_{28}}
\right).
\end{split}
\end{equation}

\subsection*{Auxiliary flux terms}

The gross canopy photosynthesis rate is
\begin{equation}
\begin{split}
\phi_{\mathrm{phot},c}(t)
=
&\frac{
\left(1-\exp\left(-p_3x_1(t)\right)\right)p_4 d_1(t)
}{
\varphi(t)
} \\
&\times
\left(-p_5x_3^2(t)+p_6x_3(t)-p_7\right)
\left(x_2(t)-p_8\right).
\end{split}
\end{equation}

The auxiliary denominator is defined as
\begin{equation}
\begin{split}
\varphi(t)
=
&p_4d_1(t) \\
&+
\left(-p_5x_3^2(t)+p_6x_3(t)-p_7\right)
\left(x_2(t)-p_8\right).
\end{split}
\end{equation}

The CO$_2$ exchange through ventilation is
\begin{equation}
\phi_{\mathrm{vent},c}(t)
=
\left(10^{-3}u_2(t)+p_{11}\right)
\left(x_2(t)-d_2(t)\right).
\end{equation}

The canopy transpiration term is
\begin{equation}
\begin{split}
\phi_{\mathrm{transp},h}(t)
=
&p_{21}
\left(1-\exp\left(-p_3x_1(t)\right)\right) \\
&\times
\left[
\frac{p_{22}}{p_{23}\bigl(x_3(t)+p_{24}\bigr)}
\exp\left(
\frac{p_{25}x_3(t)}{x_3(t)+p_{26}}
\right)
-
x_4(t)
\right].
\end{split}
\end{equation}

The humidity exchange through ventilation is
\begin{equation}
\phi_{\mathrm{vent},h}(t)
=
\left(10^{-3}u_2(t)+p_{11}\right)
\left(x_4(t)-d_4(t)\right).
\end{equation}

\subsection*{Model parameters}

The parameter vector is
\begin{equation}
p=
\begin{bmatrix}
p_1 & p_2 & \cdots & p_{28}
\end{bmatrix}^\top .
\end{equation}

The nominal parameter values used in the greenhouse model are listed in
Table~\ref{tab:lettuce_model_parameters}.

\begin{table*}[!t]
\centering
\caption{Nominal parameter values of the lettuce greenhouse model.}
\label{tab:lettuce_model_parameters}
\footnotesize
\setlength{\tabcolsep}{8pt}
\begin{tabular}{cccccc}
\hline
Parameter & Value & Parameter & Value & Parameter & Value \\
\hline
$p_1$  & 0.544                 & $p_{11}$ & $7.50\times10^{-6}$ & $p_{21}$ & $3.60\times10^{-3}$ \\
$p_2$  & $2.65\times10^{-7}$   & $p_{12}$ & 8.31                & $p_{22}$ & 9348 \\
$p_3$  & 53                    & $p_{13}$ & 273.15              & $p_{23}$ & 8314 \\
$p_4$  & $3.55\times10^{-9}$   & $p_{14}$ & 101325              & $p_{24}$ & 273.15 \\
$p_5$  & $5.11\times10^{-6}$   & $p_{15}$ & 0.044               & $p_{25}$ & 17.4 \\
$p_6$  & $2.30\times10^{-4}$   & $p_{16}$ & $3.00\times10^{4}$  & $p_{26}$ & 239 \\
$p_7$  & $6.29\times10^{-4}$   & $p_{17}$ & 1290                & $p_{27}$ & 17.269 \\
$p_8$  & $5.20\times10^{-5}$   & $p_{18}$ & 6.1                 & $p_{28}$ & 238.3 \\
$p_9$  & 4.1                   & $p_{19}$ & 0.2                 &          &       \\
$p_{10}$ & $4.87\times10^{-7}$ & $p_{20}$ & 4.1                 &          &       \\
\hline
\end{tabular}
\end{table*}

\subsection*{Remark on discretization}

For controller design, the above continuous-time model is discretized into the
nonlinear state-space form
\begin{equation}
\begin{aligned}
x(k+1) &= f\bigl(x(k),u(k),d(k),p\bigr),\\
y(k) &= g\bigl(x(k),p\bigr),
\end{aligned}
\end{equation}
using a fourth-order Runge--Kutta scheme with sampling interval
$\Delta t = 30$ min.

\bibliographystyle{unsrtnat}
\bibliography{myref}
\end{document}